%% file: main.tex
\pdfoutput=1

\documentclass[conference]{IEEEtran}
\IEEEoverridecommandlockouts

\usepackage[utf8]{inputenc}

\usepackage[style=ieee,
            citestyle=numeric-comp,
            terseinits=true]{biblatex}
\usepackage{newtxtext}

\usepackage[caption=false]{subfig}

\usepackage[utf8]{inputenc} 
\usepackage[T1]{fontenc}    
\usepackage{hyperref}       
\usepackage{url}            
\usepackage{booktabs}       
\usepackage{amsfonts}       
\usepackage{nicefrac}       
\usepackage{microtype}      
\usepackage{xcolor}         
\usepackage{graphicx}
\usepackage{wrapfig}

\usepackage{amsmath}
\usepackage{amssymb}
\usepackage{mathtools}
\usepackage{amsthm}
\usepackage{float}
\usepackage{algorithm}
\usepackage{algpseudocode}
\usepackage{multirow}
\usepackage{listings}
\usepackage{ulem}

\newcommand{\insquad}{\textsc{InSQuaD}}

\usepackage{color, colortbl}
\definecolor{LightCyan}{rgb}{0.88,1,1}
\definecolor{LightGreen}{rgb}{0.56, 0.93, 0.56}

\def\BibTeX{{\rm B\kern-.05em{\sc i\kern-.025em b}\kern-.08em
    T\kern-.1667em\lower.7ex\hbox{E}\kern-.125emX}}
\begin{document}

\title{\insquad: In-Context Learning for Efficient Retrieval via Submodular Mutual Information to Enforce Quality and Diversity}

\author{
\IEEEauthorblockN{Souradeep Nanda\textsuperscript{*}\thanks{*Equal Contribution.}}
\IEEEauthorblockA{\textit{Computer Science} \\
\textit{The University of Texas at Dallas}\\
Dallas, USA \\
souradeep.nanda@utdallas.edu}
\and
\IEEEauthorblockN{Anay Majee\textsuperscript{*}}
\IEEEauthorblockA{\textit{Computer Science} \\
\textit{The University of Texas at Dallas}\\
Dallas, USA \\
anay.majee@utdallas.edu}
\and
\IEEEauthorblockN{Rishabh Iyer}
\IEEEauthorblockA{\textit{Computer Science} \\
\textit{The University of Texas at Dallas}\\
Dallas, USA \\
rishabh.iyer@utdallas.edu}
}

\maketitle

\begin{abstract}
In this paper, we introduce, \textbf{\insquad}\footnote{Code Released at \url{https://github.com/Ghost---Shadow/InSQuaD}.}, designed to enhance the performance of \textit{\textbf{In}-Context Learning} (ICL) models through \textit{\textbf{S}ubmodular Mutual Information} (SMI) enforcing \textit{\textbf{Qua}lity} and \textit{\textbf{D}iversity} among in-context exemplars. \insquad\ achieves this through two principal strategies:
First, we model the ICL task as a targeted selection problem and introduce a unified selection strategy based on SMIs which mines relevant yet diverse in-context examples encapsulating the notions of quality and diversity. 
Secondly, we address a common pitfall in existing retrieval models which model query relevance, often overlooking diversity, critical for ICL. 
\insquad\ introduces a combinatorial training paradigm which learns the parameters of an SMI function to enforce both quality and diversity in the retrieval model through a novel likelihood-based loss.
To further aid the learning process we augment an existing multi-hop question answering dataset with synthetically generated paraphrases. 
Adopting the retrieval model trained using this strategy alongside the novel targeted selection formulation for ICL on nine benchmark datasets shows significant improvements validating the efficacy of our approach.\looseness-1
\end{abstract}


\input{10_introduction}

\input{20_related_work}

\input{30_methodology}

\input{50_results}

\input{60_ablations}

\input{70_outro}

\printbibliography

\end{document}

%% file: 10_introduction.tex
\section{Introduction}
In-Context Learning (ICL)~\cite{Zhang2022ActiveES, Wu2022SelfAdaptiveIL, Li2023FindingSE} has soared to prominence in NLP by cutting down on task-specific training and enabling swift generalization to downstream tasks. It accomplishes this by injecting \textit{few-shot} prompts into the context of the LLM, hereby known as \textit{exemplars}, or \textit{demonstrations}—to guide a Large Language Model (LLM) at test time. Following prior works~\cite{Kumari2024AnES, Zhang2023IDEALIS, Su2022SelectiveAM}, selecting these exemplars unfolds in two stages. First, we shortlist diverse exemplars from a large unlabeled corpus for human annotation (\textit{Exemplar Annotation}). Next, we retrieve a small subset from this newly annotated pool that is highly relevant to the input query (\textit{Exemplar Retrieval}). Finally, this subset, bundled with the query, is presented to a pretrained LLM, which produces the desired solution.\looseness-1

Several recent works~\cite{Su2022SelectiveAM,Wu2022SelfAdaptiveIL} underscore that the success of ICL hinges on the \textit{quality} and \textit{diversity} of in-context exemplars and employ selective annotation~\cite{Mavromatis2023WhichET} to lighten the burden of labeling massive datasets by focusing on query relevance (\textit{quality}), oftentimes neglecting diversity among selected exemplars. 
Consequently, several works adopt combinatorial techniques~\cite{Zhang2023IDEALIS, Kumari2024AnES} modeling ICL as a selection/summarization task, identifying prompts most pertinent to the input query. However, all selection techniques operate on a constant budget, risking the exclusion of diverse yet relevant in-context examples. Meanwhile, techniques~\cite{Ye2023CompositionalEF, Rubin2021LearningTR} that learn the parameters of the underlying retrieval model prioritize query relevance but frequently overlook an explicit treatment of both quality and diversity in the learned model.\looseness-1

\input{figures/title_figure}

To this end, we introduce \textbf{\insquad}, a unified combinatorial approach for \textbf{In}-Context Learning. Drawing on \textbf{S}ubmodular Mutual Information (SMI) functions~\cite{Iyer2022GeneralizedSI,Kothawade2021PRISMAR}, \insquad\ naturally incorporates \textbf{Qua}lity, \textbf{D}iversity, and Order among exemplars in two phases: \insquad-RETRIEVE and \insquad-LEARN as shown in Figure \ref{fig:insquad_title_fig}. 
First, \insquad-RETRIEVE views selective annotation as a targeted selection problem applied in two distinct stages as discussed above.
During Exemplar Annotation, diversity is enforced by maximizing the SMI among an unlabeled pool of exemplars, identifying distinct examples for annotation. 
Exemplar Retrieval extracts the most relevant yet diverse examples for answering an input query at runtime. Unlike existing works which either model query relevance~\cite{Lewis1994ASA, Rubin2021LearningTR, Su2022SelectiveAM} or employ a multi-stage approach~\cite{Kumari2024AnES} to model quality and diversity, our SMI based targeted selection strategy jointly models both traits while selecting in-context exemplars.
Further, adopting greedy optimization~\cite{Nemhauser1978AnAO} in maximizing SMI models an implicit ordering of exemplars by their incremental submodular information gains. 
Thus, \insquad\ elegantly weaves together all three pillars: quality, diversity, and order.\looseness-1

Secondly, we note that current methods rely on embeddings from pretrained retrieval models~\cite{Reimers2019SentenceBERTSE} that prioritize query relevance but \textit{overlook diversity}. 
\insquad-LEARN addresses this gap by introducing a family of likelihood-based loss functions (Table \ref{tab:instances_quaild}) to train a retrieval model that \textbf{learns the parameters of an SMI function}, leveraging the property of submodular functions which captures cooperation~\cite{Jegelka2011SubmodularityBS} and diversity~\cite{Lin2011ACO} when minimized and maximized respectively.
This formulation thus aligns with the goal of the downstream ICL tasks which is represented as a selection problem in \insquad.
Finally, we facilitate the training of underlying retrieval models by introducing a novel dataset, augmenting exisitng multi-hop question answering datasets like HotpotQA~\cite{Yang2018HotpotQAAD} with synthetic documents (Section \ref{sec:insquad_learn}) aiding the capture of quality (contextual similarity) and diversity (paraphrased distractors) in the learnt representations.\looseness-1

The main contributions of \insquad\ can be summarized as follows - 
\begin{itemize}
    \item \insquad\ introduces a novel \textbf{combinatorial targeted selection strategy in selecting in-context prompts for ICL based on SMI functions}, inherently modeling the notions of \textit{quality} and \textit{diversity}.\looseness-1

    \item \insquad\ introduces a novel training paradigm for fine-tuning existing sentence embedding models to inherently model quality and diversity, by \textbf{learning the parameters of a likelihood based SMI objective}, a necessity for downstream ICL tasks.\looseness-1

    \item Applying both \insquad-RETRIEVE and \insquad-LEARN shows improvements of up to 21.6\% on classification tasks, 16.4\% on multi-choice tasks and up to 7\% on generation based ICL tasks (refer Sec.\ref{sec:main_results}) over existing baselines.\looseness-1
\end{itemize}

%% file: figures/title_figure.tex
\begin{figure*}[t]
  \centering
  \includegraphics[width=\textwidth]{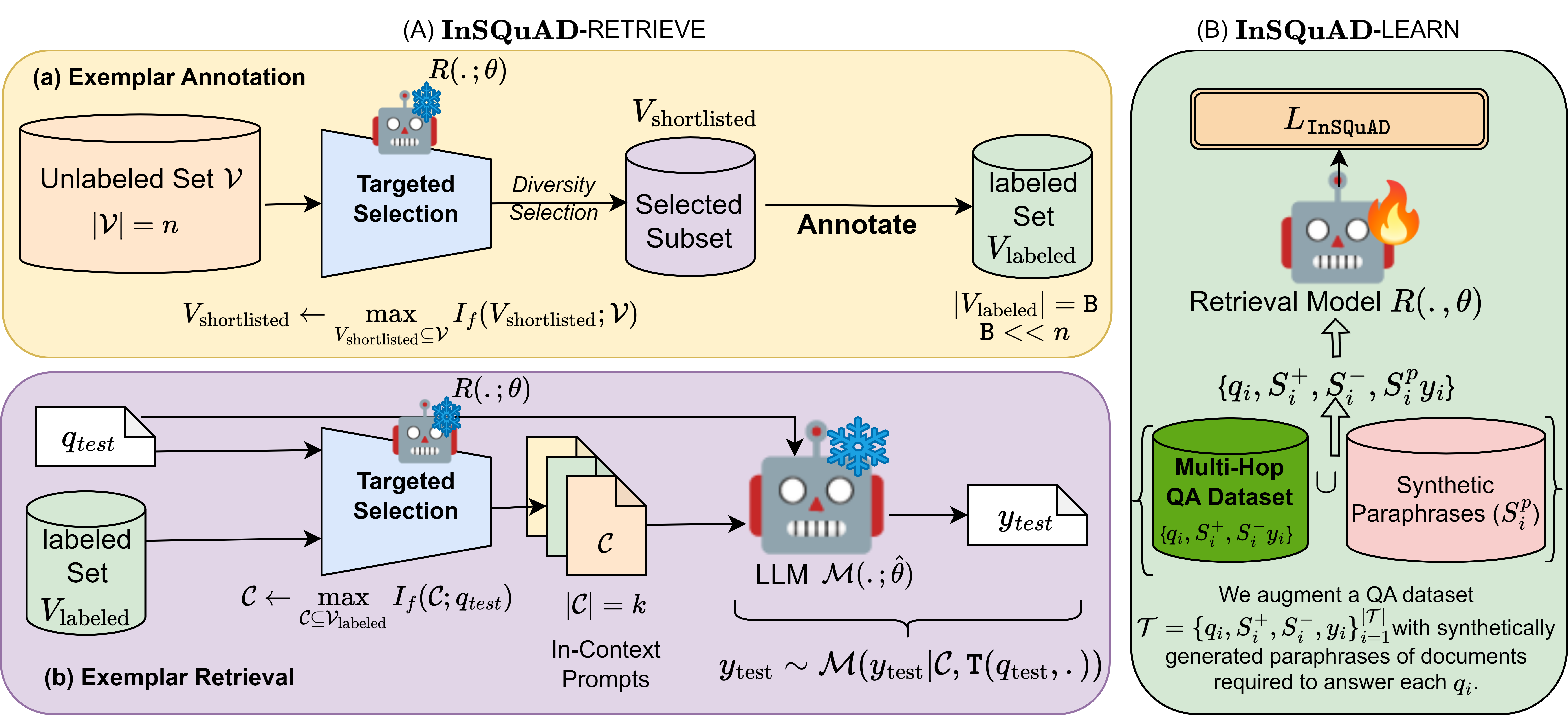}
  \caption{\textbf{Overview of \insquad} comprising two principle components - (a) \textbf{\insquad-RETRIEVE} which models exemplar annotation and retrieval as a targeted selection problem injecting the notions of \textit{quality}, \textit{diversity} and \textit{order}. (b) \textbf{\insquad-LEARN} which introduces a novel learning strategy to enforce the notions of quality and diversity into the retrieval model $R$.}
  \label{fig:insquad_title_fig}
\end{figure*}

%% file: 20_related_work.tex

\section{Related Work}
\label{sec:rel_work}

\noindent \textbf{In-Context Learning:} 
The exploration of in-context learning has significantly advanced our understanding of large language model (LLM) capabilities. Active example selection has been highlighted as a crucial factor for enhancing in-context learning, with \cite{Zhang2022ActiveES}, \cite{Wu2022SelfAdaptiveIL}, \cite{Diao2023ActivePW} and \cite{Li2023FindingSE} providing insights into the process. The role of demonstrations in in-context learning effectiveness has been examined by \cite{Min2022RethinkingTR} and  \cite{Kim2022GroundTruthLM}, emphasizing the impact of demonstration quality. Theoretical underpinnings of in-context learning are proposed by \cite{Xie2021AnEO} and \cite{Hahn2023ATO}, discussing mechanisms like implicit Bayesian inference and structure induction. Research by \cite{Li2023UnifiedDR}, \cite{Han2023UnderstandingIL}, \cite{Bansal2022RethinkingTR} and  \cite{Wei2023LargerLM} explores the scalability of in-context learning, while the function of induction heads in this paradigm is investigated by \cite{Olsson2022IncontextLA}.\looseness-1


\noindent \textbf{Language Model and Learning Strategies:}
The combination of active learning strategies with transformer architecture has paved the way for significant progress in the field. The work of \cite{EinDor2020ActiveLF} and \cite{Schrder2021RevisitingUQ} exemplify advancements in active learning for transformers. The seminal paper by \cite{Vaswani2017AttentionIA}, introducing the transformer model, marks a cornerstone in language model research. Further exploration into the learning mechanisms of transformers is provided by \cite{Oswald2022TransformersLI} and \cite{Zhang2023TrainedTL}. Benchmark leakage analysis done by \cite{Zhou2023DontMY} sheds interesting insights. The broader implications of foundation models are discussed by \cite{Bommasani2021OnTO}, complemented by surveys and studies on the emergent abilities and challenges in scaling language models \cite{Zhao2023ASO} \cite{Dong2022ASO,Wei2022EmergentAO}. Lastly, the importance of data distributional properties in emergent in-context learning is highlighted by \cite{Chan2022DataDP}.\looseness-1

%% file: 30_methodology.tex
\section{Method}
\label{sec:method}

\subsection{Problem Formulation and Notations}
Given a search query $q_{\text{test}}$ and a set of relevant (in-context) demonstrations $\mathcal{C}$, an In-Context Learner (ICL) aims to generate a solution $y_{\text{test}}$ from an LLM $\mathcal{M}(.; \hat{\theta})$, parameterized by $\hat{\theta}$ as:
\begin{align}
    p(y_{\text{test}}|\mathcal{C}, q_{\text{test}}) = \mathcal{M}(\mathtt{V}(y_{\text{test}}) | \mathcal{C}, \mathtt{T}(q_{\text{test}}, .);\hat{\theta})
\end{align}
The in-context demonstrations (also known as exemplars) in $\mathcal{C} = \{\mathtt{T}(q_i, y_i)\}_{i = 1}^{k}$ is mined from an unlabeled pool of documents $\mathcal{V}$ in a few-shot ($k << |\mathcal{V}|$) fashion, where $\mathcal{C} \subseteq \mathcal{V}$ based on their relevance to the test query $q_{\text{test}}$.
Here, the query and each in-context exemplar in $\mathcal{C}$ is templated using a common template indicated by $\mathtt{T}(.)$ while the output $y_{\text{test}}$ is verbalized using the verbalizer $\mathtt{V}(.)$ as defined in \cite{Ye2023CompositionalEF}.
The generation of $y_{\text{test}}$ proceeds without updating the parameters of the LLM $\mathcal{M}(., \hat{\theta})$, solely depending on the relevance and diversity of the in-context demonstrations to $q_{\text{test}}$.\looseness-1 

\subsection{Preliminaries: Submodularity}
\label{sec:submod_prelims}
Submodular functions are set functions characterized by a natural diminishing returns property. Specifically, a set function \( f: 2^\mathcal{V} \rightarrow \mathbb{R} \), defined over a ground set \(\mathcal{V}\), is submodular if it satisfies the inequality \( f(A_i) + f(A_j) \geq f(A_i \cup A_j) + f(A_i \cap A_j) \) for all subsets \( A_i, A_j \subseteq \mathcal{V} \)~\cite{1991SubmodularFA}. 
These functions have been extensively studied for data subset selection~\cite{Killamsetty2022AUTOMATAGB, Kothawade2021PRISMAR, Kothawade2021TALISMANTA, Jain2024EfficientDS}, video summarization~\cite{Kaushal2019DemystifyingMV, Kaushal2021HowGI} etc. 
Traditionally, these tasks are modeled as a discrete optimization problem through submodular maximization under a knapsack constraint~\cite{Nemhauser1978AnAO}. This can be fairly approximated with a $(1 - e^{-1})$ constant factor guarantee using greedy optimization techniques~\cite{Mirzasoleiman2014LazierTL}.
In our paper we study two popular categories of submodular functions - \textbf{Submodular Information Functions} (SIMs) and \textbf{Submodular Mutual Information Functions} (SMIs). 
Maximizing a SIM~\cite{1991SubmodularFA} $f(A)$ like Facility-Location, Graph-Cut etc. promotes selection of diverse examples within a set $A$, while maximizing a SMI function $I_f(A;Q)$ (given the underlying submodular function $f$) selects examples that share maximum information in $A \cap Q$. 
A special class of submodular functions, Submodular Point Processes (SPPs) \cite{Iyer2015SubmodularPP} presents a probabilistic likelihood formulation which has been used to learn the parameters of an underlying combinatorial function for summarization tasks.
Very recently combinatorial functions have been applied as learning objectives in continuous optimization problems like longtail recognition~\cite{Majee2023SCoReSC} and few-shot object detection~\cite{Majee2024SMILeLS} for representation learning. 
Although several ICL approaches (refer Sec.\ref{sec:rel_work}) adopt combinatorial approaches, we present a unified method incubating all three notions of \textit{quality} and \textit{diversity}, a known pitfall in literature.\looseness-1

\subsection{\insquad}
\label{sec:insquad}
In this section we outline the details of our proposed \insquad\ method that unifies the notions of \textit{quality} and \textit{diversity} into a novel combinatorial formulation. 
\insquad\ achieves this in two distinct stages - \textbf{(1) \insquad-RETRIEVE} which performs both exemplar annotation and retrieval to shortlist a small yet diverse set of exemplars from an unlabeled pool (exemplar annotation) and selecting the most relevant yet diverse exemplars (exemplar retrieval) for downstream ICL tasks.
In contrast to other approaches in literature (refer Sec. \ref{sec:rel_work}) which adopt separate functions for each of the above tasks \insquad\ solely relies on Submodular Mutual Information (SMI) functions modeling both tasks as a targeted selection problem discussed in Sec. \ref{sec:insquad_retrieve}. The embedding generator used during selection is a retrieval model $R(., \theta)$ trained using 
\textbf{(2) \insquad-LEARN} which learns its parameters $\theta$ to encapsulate both the notions of quality and diversity, unlike popularly used models~\cite{Reimers2019SentenceBERTSE} which inculcate only query relevance. 
\insquad\ achieves this through a novel combinatorial loss formulation, also based on SMI functions elucidated in Sec. \ref{sec:insquad_learn}. The trained weights of $R$ is consumed directly by \insquad-RETRIEVE which serves as an embedding generator during both the annotation and retrieval stages (weights in $R$ are frozen).

\looseness-1

\begin{algorithm}[t]
    \caption{Exemplar Annotation and Retrieval in \insquad}
    \label{alg:inference}
    \begin{algorithmic}[1]
    \Require Ground Set $V$, Query Set $Q$, selection budget $k$.
    \Procedure{\textsc{TargetedSelection}}{$Q, V, k$}
        \State $S \gets \emptyset$ \Comment{Initialize the selected subset}
        \While{$|S| < k$} 
            \State $v^* \gets \arg\max_{v \in V \setminus S} I_f(S \cup \{v\}; Q) - I_f(S; Q)$ 
            \State $S \gets S \cup \{v^*\}$ 
        \EndWhile
        \State \textbf{return} $S$ 
    \EndProcedure
    \end{algorithmic}
\end{algorithm}
        
\subsubsection{\insquad-RETRIEVE}
\label{sec:insquad_retrieve}
A core task in ICL is to mine diverse yet relevant in-context examples, required to answer $q_{\text{test}}$. 
We model this task as a two-stage targeted selection problem with the introduction of SMI Functions ($I_f$), namely \textit{Exemplar Annotation} and \textit{Exemplar Retrieval}.
The \textbf{Exemplar Annotation} stage circumvents the challenge of annotating the complete unlabeled set $\mathcal{V}$ by selecting a diverse yet representative subset of examples $V_{\text{shorlisted}} \subseteq \mathcal{V}$. We achieve this by maximizing the SMI over the unlabeled set as shown in eq. \ref{eq:smi_shortlisting}.\looseness-1
\begin{align}
    V_{\text{shortlisted}}  \leftarrow \underset{\substack{V_{\text{shortlisted}} \subseteq \mathcal{V} \\ |V_{\text{shortlisted}}| \leq \mathtt{B}}}{\mathtt{argmax}} I_f(V_{\text{shortlisted}} ; \mathcal{V})
    \label{eq:smi_shortlisting}
\end{align}
Here, $\mathtt{B}$ is the annotation budget $\mathtt{B} \leq |\mathcal{\mathcal{V}}|$. 
Note that since the query set in the SMI function $I_f$ is the complete ground set $\mathcal{V}$, this formulation boils down to maximizing the submodular function $f$ over the unlabeled set $\mathcal{V}$ as $V_{\text{shortlisted}} \leftarrow \underset{\substack{V_{\text{shortlisted}} \subseteq \mathcal{V} \text{ , } |V_{\text{shortlisted}}| \leq \mathtt{B}}}{\mathtt{argmax}} f(V_{\text{shortlisted}})$ as shown in \cite{Iyer2020SubmodularCI}.
Following the observations in \cite{Lin2012LearningMO} \textit{maximizing the submodular information $f$ over a subset $V_{\text{shortlisted}}$ models diversity} among selected examples agnostic of the test query.
Exemplars in $V_{\text{shortlisted}}$ are further labeled by human annotators to produce a labeled set $V_{\text{labeled}}$.\looseness-1

\input{figures/insquad_learn}

Given a test query $q_{\text{test}}$ and the labeled set of exemplars $V_{\text{labeled}}$, the \textbf{Exemplar Retrieval} phase selects the top-$k$ in-context exemplars $\mathcal{C}$, most relevant to answering $q_{\text{test}}$.
Although previous research (refer Sec.\ref{sec:rel_work}) emphasizes the modeling of both quality and diversity among exemplars in $\mathcal{C}$, recent methods like \cite{Wu2022SelfAdaptiveIL, Su2022SelectiveAM} focus mainly on query relevance while \cite{Kumari2024AnES, Zhang2023IDEALIS} employ multi-stage selection/summarization strategies, where each stage models either relevance or diversity.
\insquad\ drifts away from existing works by introducing a unified single stage formulation as in eq. \ref{eq:smi_retrieval} by maximizing the SMI between $q_{\text{test}}$ and selected exemplars in $\mathcal{C}$, modeling the retrieval of in-context exemplars as a targeted selection problem. 

\input{tables/instances_insquad}

\begin{align}
    \mathcal{C}  \leftarrow \underset{\substack{\mathcal{C} \subseteq V_{\text{labeled}} \text{ , } |\mathcal{C}| \leq k}}{\mathtt{argmax}} \text{  } I_f(\mathcal{C}; q_{\text{test}})
    \label{eq:smi_retrieval}
\end{align}
The greedy optimization~\cite{Nemhauser1978AnAO} strategy detailed in Alg.\ref{alg:inference}, adopted during the selection process orders the exemplars in $\mathcal{C}$ based on decreasing information gain. 
This results in an inherent ordering of the exemplars based on their relevance towards answering $q_{\text{test}}$, a necessity in ICL~\cite{Lu2021FantasticallyOP}.
Note that the Exemplar Annotation phase is performed only once whereas the Exemplar Retrieval is invoked for each test query. Our results in Tab.\ref{table:main_table} show that the application of \insquad-RETRIEVE alone (indicated as NT) boosts performance over training free ICL methods showing the effectiveness of modeling all three notions of quality, diversity and order.

\begin{table*}
\centering
\footnotesize

\input{tables/main_table_gemma_ci}

\end{table*}

\subsubsection{\insquad-LEARN}
\label{sec:insquad_learn}
\label{subsec:loss_function_for_quality}
\label{subsec:loss_function_for_diversity}

Although training free strategies are largely popular in ICL to leverage knowledge from in-context exemplars, \cite{Ye2023CompositionalEF} highlights the common fallacy in LLMs~\cite{Reimers2019SentenceBERTSE} which for retrieval tasks model query relevance without incorporating the notion of diversity. 
Unlike existing approaches (refer Sec.\ref{sec:rel_work}) \insquad\ explicitly models the notions of quality and diversity into the retrieval model $R(., \theta)$, mimicking the behavior expected during downstream ICL tasks.\looseness-1

\textit{\insquad\ achieves this by modeling the modeling the learning problem of $R$ as a Submodular Point Process (SPP)}~\cite{Iyer2015SubmodularPP}. 
Given a subset $S$ of documents retrieved from a ground set $V$ and a query $Q$, SPPs estimate the $P_{\theta}^Q(S)$ of the retrieval set $S$ being relevant to answering $Q$ over all possible subsets in $V$ as shown in eq. \ref{eq:p_theta_q}.
Adopting a similar choice of information function as in \insquad-RETRIEVE we adopt SMI functions (with underlying submodular function $f$) to model shared information between $S$ and $Q$.
\begin{align}
    \label{eq:p_theta_q}
    P_{\theta}^Q(S) = \frac{I_f(S;Q)}{\sum_{S' \subset V} I_f(S';Q)}
\end{align}
To incorporate SPPs defined above into a learning objective in \insquad-LEARN given a predetermined set of relevant documents $S^+$ and distractors $S^-$, we define $\alpha_Q^\theta$ as the ratio of probability of selecting $S^+$ over distractors in $S^-$, given the parameters $\theta$ in $R$. 
\begin{align*}
\alpha_Q^\theta = \frac{P_{\theta}^Q(S^+)}{P_{\theta}^Q(S^-)} = \frac{\frac{I_f(S^+;Q)}{\sum_{S' \subset V} I_f(S';Q)}}{\frac{I_f(S^-;Q)}{\sum_{S' \subset V} I_f(S';Q)}} = \frac{I_f(S^+;Q)}{I_f(S^-;Q)}
\end{align*}
Our learning objective in \insquad-LEARN maximizes the likelihood of $S^+$ to be similar to $Q$ while minimizing its similarity to $S^-$. Thus, by taking the negative logarithm of $\alpha_Q^\theta$ we define a learning objective $L$ as shown in eq. \ref{eq:l_def} which is effectively minimized during model training.
\begin{align}
L = -\log(\alpha_Q^\theta) = \log(I_f(S^-;Q)) - \log(I_f(S^+;Q))
\label{eq:l_def}
\end{align}

Given, this loss formulation inspired from SPPs we introduce two novel formulations one enforcing \textit{quality} while the other enforcing \textit{diversity} in the training loop of $R$. Note, that $\alpha_Q^\theta$ depends on $I_f$ which in turn computes embedding similarity between $Q$ and $S^+$/$S^-$ that depends on the parameter $\theta$ of $R$ during training.
  

\noindent \textbf{Training Loop:} For every query $q_i$, $\forall i \in |\mathcal{T}|$ in the training dataset $\mathcal{T}$ (in practice $\mathcal{T}$ refers to the mini-batch) posed at the retrieval model $R(., \theta)$, we provide a set of exemplars $S_i^+$ relevant to answering $q_i$ and a set of distractors $S_i^-$, where $S_i^- \cap S_i^+ = \phi$.
Given the loss formulation in eq.\ref{eq:l_def}, \insquad-LEARN enforces quality in $R(., \theta)$ through $L_q$ (aggregated over the full dataset $\mathcal{T}$) which maximizes the feature overlap between the query $q_i$ and $S_i^+$ (Term 2 in eq.\ref{eq:quality_loss}) while minimizing the common information between $S_i^-$ and $q_i$ (Term 1 in eq.\ref{eq:quality_loss}). 

\begin{align}
	\label{eq:quality_loss}
	L_q = \sum_{i \in \mathcal{T}} \underbrace{ \log(I_f(S_i^-;q_i))}_\text{Term 1} - \underbrace{\log(I_f(S_i^+;q_i))}_\text{Term 2}
\end{align}

Although $L_q$ enforces $R(., \theta)$ to model query relevance (notion of quality), $R(., \theta)$ continues to be susceptible to paraphrases and irrelevant documents that convey similar meaning to the exemplars in $S_i^+$, degrading downstream performance. 
Given a set $S^p_i$ which are paraphrases of elements in $S_i^+$ \insquad-LEARN reuses the newly introduced loss formulation in eq.\ref{eq:l_def} to enforce a diversity based objective $L_d$ to evade retrieval of such paraphrases when deployed in downstream ICL tasks. This loss formulation, shown in eq. \ref{eq:diversity_loss} minimizes the SMI between the $S_i^-$ and the paraphrases of relevant documents $S_i^p$ (Term 1 in eq. \ref{eq:diversity_loss}) while maximizing the information overlap between the true paraphrases $S_i^p$ and the relevant documents $S_i^+$. 
\begin{equation}
    \label{eq:diversity_loss}
    L_d = \sum_{i \in \mathcal{T}}\underbrace{\log(I_f(S^-_i; S_i^p))}_\text{Term 1} - \underbrace{\log(I_f(S^+_i; S_i^p))}_\text{Term 2}
\end{equation}


Finally, \insquad\ combines the notions of quality and diversity into a joint loss formulation $L_{\insquad} = \exp((1 - \lambda)L_q + \lambda L_d)$ as shown in Fig.\ref{fig:fig_training_flow}. Here, $\lambda$ controls the quality-diversity trade-off during training with supporting experimental results in Sec. \ref{sec:quality_div_tradeoff}.
Note that by varying the choice of submodular function $f$ in both $L_q$ and $L_d$ among Graph-Cut (GC), Facility-Location (FL) and Log-Determinant (LD) we create three novel instances of $L_{\insquad}$ as summarized in Tab. \ref{tab:instances_quaild}.
Results from our experiments in Tab. \ref{table:main_table} combines \insquad-LEARN and \insquad-RETRIEVE into a unified formulation showing significant performance boosts over baseline methods. Note, that all instances depend on computation of a similarity kernel $S_{ab}$ to encode interaction between documents $a$ and $b$ as in \cite{Majee2023SCoReSC}.\looseness-1


\input{figures/ablations}

\noindent \textbf{Training Data Curation:} The training loop in \insquad\ heavily relies on irrelevant documents acting as distractors and paraphrases alongside relevant ones to bake the notions of quality and diversity into the training regime.
Unfortunately popular multi-hop Question Answering (QA) datasets~\cite{Rajpurkar2016SQuAD1Q, Yang2018HotpotQAAD} do not encapsulate paraphrases, thereby hindering models trained on them to model de-duplication impacting diversity (critical for ICL).
To this end, given a standard QA dataset $\mathcal{T}$ we augment it with synthetically generated paraphrases to create our training data. 
Each row in the multihop QA dataset is of form $\{q_i, S_i^+, S_i^-, y_i\}$ where $q_i$ is the question, $V_{raw} = S_i^+ \cup S_i^-$ are all supporting documents, $S^+$ being relevant documents and $S_i^-$ are the distractors for retrieving the answer $y_i$. We synthetically generate one paraphrase for each item in $V_{raw}$ and call the set $S_i^p$.
We leverage GPT-3.5 Turbo~\cite{Ouyang2022TrainingLM} to generate paraphrases for each corresponding document in $V_{raw}$ for each question $q_i$. 
This follows the trend in existing works~\cite{Sachan2022QuestionsAA,Li2022AskTU} which generate synthetic questions/documents for auxiliary NLP tasks.
Although \textit{any} multi-hop QA dataset can be used here, our experiments adopt the HotpotQA~\cite{Yang2018HotpotQAAD} dataset owing to its popularity in in multi-hop QA reflected through high citation volume (2349 citations) over contemporary multi-hop question answering datasets like MuSiQue~\cite{trivedi2021musique} (207 citations), ConcurrentQA~\cite{arora2022reasoning} (22 citations). We will release the augmented dataset publicly upon review.\looseness-1

%% file: figures/insquad_learn.tex
\begin{figure}[t]
  \centering
  \includegraphics[width=\columnwidth]{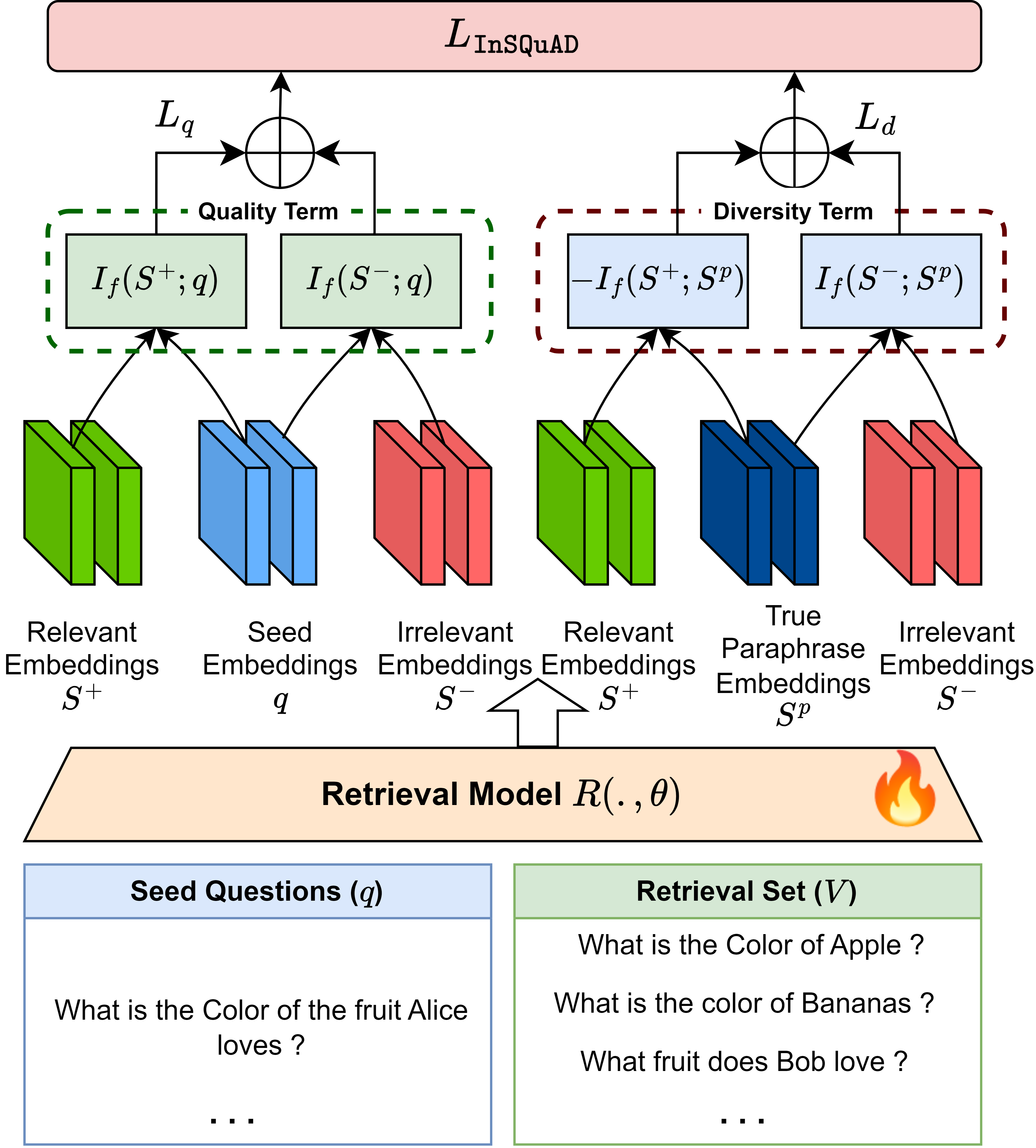}
  \caption{\textbf{Training Workflow of \insquad-LEARN} which learns the parameters of an SMI function to update the parameters of a retrieval model $R$ through a novel likelihood based objective $L_{\insquad}$ enforcing quality and diversity in learnt embeddings.}
  \label{fig:fig_training_flow}
\end{figure}

%% file: tables/instances_insquad.tex
\begin{table*}[t]
      \centering
      \small
      \renewcommand{\arraystretch}{1.5} 
      \setlength{\tabcolsep}{1pt}
        \caption{\textbf{Instantiations of \insquad-LEARN} obtained by varying the choice of the underlying submodular function $f$ among Facility-Location (\insquad-FL), Graph-Cut (\insquad-GC) and Log-Determinant (\insquad-LD) in $L_q$ and $L_d$ respectively. Here, $S_{ab}$ is the similarity kernel between document embeddings $a$ and $b$.\looseness-1}
      \label{table:submod_functions}
      \label{tab:instances_quaild}
      \resizebox{\textwidth}{!}{\begin{tabular}{l|c|c}
            \toprule
            \textbf{Instance Name}  & 
            \textbf{Instances of $L_q$ for quality} & 
            \textbf{Instances of $L_d$ for diversity} \\
            \midrule \midrule
            \insquad-GC &            
            $ \sum_{i \in \mathcal{T}} \log\left( \frac{\sum_{a \in S_i^-} \sum_{b \in q_i} S_{ab}}{\sum_{a \in S_i^+} \sum_{b \in q_i} S_{ab}} \right)$ &
            $ \sum_{i \in \mathcal{T}} \log\left( \frac{ \sum_{a \in S^-_i} \sum_{b \in S_i^p} S_{ab} }{ \sum_{a \in S^+_i} \sum_{b \in S_i^p} S_{ab} } \right)$ \\ \midrule

            \insquad-FL &
            $\sum_{i \in \mathcal{T}} \log \left( \frac{ \sum_{a \in S_i^-} \max_{b \in q_i} S_{ab} + \eta \sum_{b \in q_i} \max_{a \in S_i^-} S_{ab} }{ \sum_{a \in S_i^+} \max_{b \in q_i} S_{ab} + \eta \sum_{b \in q_i} \max_{a \in S_i^+} S_{ab} } \right)$ &
            $\sum_{i \in \mathcal{T}} \log\left( \frac{ \sum_{a \in S^-_i} \max_{b \in S_i^p} S_{ab} + \eta \sum_{b \in S_i^p} \max_{a \in S^-_i} S_{ab} }{ \sum_{a \in S^+_i} \max_{b \in S_i^p} S_{ab} + \eta \sum_{b \in S_i^p} \max_{a \in S^+_i} S_{ab} } \right)$ \\ \midrule

            \insquad-LD &
            $\sum_{i\in \mathcal{T}} \log \frac{\log\det\bigl(S_{S_i^-}\bigr) - \log\det\Bigl(S_{S_i^-} - \eta^2\, S_{S_i^-,q_i}\, S_{q_i}^{-1}\, S_{S_i^-,q_i}^T\Bigr)}{\log\det\bigl(S_{S_i^+}\bigr) - \log\det\Bigl(S_{S_i^+} - \eta^2\, S_{S_i^+,q_i}\, S_{q_i}^{-1}\, S_{S_i^+,q_i}^T\Bigr)}$ &
            $\sum_{i \in \mathcal{T}} \log \frac{\log\det\bigl(S_{S_i^-}\bigr) - \log\det\Bigl(S_{S_i^-} - \eta^2\, S_{S_i^-,S_i^p}\, S_{S_i^p}^{-1}\, S_{S_i^-,S_i^p}^T\Bigr)}{\log\det\bigl(S_{S_i^+}\bigr) - \log\det\Bigl(S_{S_i^+} - \eta^2\, S_{S_i^+,S_i^p}\, S_{S_i^p}^{-1}\, S_{S_i^+,S_i^p}^T\Bigr)} $ \\
            \bottomrule
      \end{tabular}}\\
\end{table*}

%% file: tables/main_table_gemma_ci.tex
\caption{\textbf{Performance of our INSQUAD against existing approaches}, evaluated across nine distinct datasets on Gemma (2B). Our approach outperforms existing baselines on retrieval with the top-performing result for each dataset is highlighted in \textbf{bold}.}
\label{table:main_table}

\setlength{\tabcolsep}{3pt}
\begin{tabular}{@{}l|ccccccccc|cc@{}}
\toprule
Method & \multicolumn{5}{c}{\textbf{Classification}} & \textbf{Multi-Choice} & \textbf{Dialogue} & \multicolumn{2}{c|}{\textbf{Generation}} & \textbf{Avg.} & \textbf{Avg.} \\
\cmidrule(lr){2-10}
 & \textsc{MRPC} & \textsc{SST5} & \textsc{MNLI} & \textsc{DBpedia} & \textsc{RTE} & \textsc{HellaSwag} & \textsc{MWoZ} & \textsc{GeoQ} & \textsc{Xsum} & \textbf{Rank} & \textbf{Perf.} \\
\midrule \midrule
Zeroshot & 0.28$_{\pm0.02}$ & 0.26$_{\pm0.02}$ & 0.41$_{\pm0.01}$ & 0.52$_{\pm0.02}$ & 0.55$_{\pm0.02}$ & 0.19$_{\pm0.01}$ & 0.07$_{\pm0.00}$ & 0.63$_{\pm0.01}$ & 0.18$_{\pm0.00}$ & 12.2 & 0.34 \\
Random & 0.51$_{\pm0.02}$ & 0.42$_{\pm0.01}$ & 0.37$_{\pm0.02}$ & 0.57$_{\pm0.00}$ & 0.52$_{\pm0.04}$ & 0.27$_{\pm0.01}$ & 0.27$_{\pm0.01}$ & 0.82$_{\pm0.09}$ & 0.23$_{\pm0.02}$ & 8.9 & 0.44 \\
Diversity & 0.45$_{\pm0.02}$ & 0.42$_{\pm0.03}$ & 0.39$_{\pm0.03}$ & 0.59$_{\pm0.04}$ & 0.55$_{\pm0.03}$ & 0.24$_{\pm0.01}$ & 0.16$_{\pm0.01}$ & 0.89$_{\pm0.03}$ & 0.23$_{\pm0.00}$ & 8.6 & 0.44 \\
Least Confidence & 0.56$_{\pm0.03}$ & 0.33$_{\pm0.02}$ & 0.32$_{\pm0.02}$ & 0.45$_{\pm0.00}$ & 0.57$_{\pm0.04}$ & 0.27$_{\pm0.02}$ & 0.13$_{\pm0.00}$ & 0.86$_{\pm0.01}$ & 0.23$_{\pm0.01}$ & 9.4 & 0.41 \\
MFL & 0.51$_{\pm0.01}$ & 0.43$_{\pm0.02}$ & 0.32$_{\pm0.01}$ & 0.62$_{\pm0.02}$ & 0.52$_{\pm0.02}$ & 0.29$_{\pm0.01}$ & 0.39$_{\pm0.04}$ & 0.89$_{\pm0.03}$ & 0.12$_{\pm0.00}$ & 7.6 & 0.46 \\
GC & 0.47$_{\pm0.01}$ & 0.38$_{\pm0.02}$ & 0.35$_{\pm0.01}$ & 0.60$_{\pm0.01}$ & 0.51$_{\pm0.03}$ & 0.21$_{\pm0.01}$ & 0.18$_{\pm0.01}$ & 0.88$_{\pm0.01}$ & 0.24$_{\pm0.00}$ & 10.8 & 0.42 \\
Vote-K & 0.47$_{\pm0.01}$ & 0.40$_{\pm0.01}$ & 0.33$_{\pm0.02}$ & 0.63$_{\pm0.01}$ & 0.52$_{\pm0.01}$ & 0.25$_{\pm0.02}$ & 0.33$_{\pm0.02}$ & 0.89$_{\pm0.02}$ & 0.18$_{\pm0.01}$ & 9.9 & 0.44 \\
IDEAL & 0.47$_{\pm0.02}$ & 0.42$_{\pm0.01}$ & 0.35$_{\pm0.01}$ & 0.62$_{\pm0.02}$ & 0.54$_{\pm0.01}$ & 0.26$_{\pm0.00}$ & 0.36$_{\pm0.01}$ & 0.82$_{\pm0.07}$ & 0.19$_{\pm0.01}$ & 8.9 & 0.45 \\
\midrule
InSQuaD-FL (NT) & 0.50$_{\pm0.00}$ & 0.40$_{\pm0.04}$ & 0.32$_{\pm0.01}$ & 0.66$_{\pm0.05}$ & 0.52$_{\pm0.01}$ & 0.28$_{\pm0.01}$ & 0.30$_{\pm0.02}$ & 0.80$_{\pm0.04}$ & 0.10$_{\pm0.01}$ & 10.3 & 0.43 \\
InSQuaD-LD (NT) & 0.47$_{\pm0.03}$ & 0.38$_{\pm0.03}$ & 0.37$_{\pm0.01}$ & 0.68$_{\pm0.02}$ & 0.55$_{\pm0.02}$ & 0.25$_{\pm0.04}$ & 0.39$_{\pm0.00}$ & 0.83$_{\pm0.01}$ & 0.22$_{\pm0.03}$ & 8.7 & 0.46 \\
InSQuaD-GC (NT) & 0.57$_{\pm0.01}$ & 0.41$_{\pm0.03}$ & 0.34$_{\pm0.02}$ & 0.65$_{\pm0.03}$ & 0.52$_{\pm0.01}$ & 0.27$_{\pm0.02}$ & 0.34$_{\pm0.02}$ & 0.84$_{\pm0.00}$ & 0.19$_{\pm0.01}$ & 8.4 & 0.46 \\
\midrule
\rowcolor{LightGreen}
InSQuaD-FL & 0.50$_{\pm0.02}$ & 0.42$_{\pm0.02}$ & 0.39$_{\pm0.02}$ & 0.65$_{\pm0.01}$ & 0.56$_{\pm0.00}$ & 0.27$_{\pm0.02}$ & 0.34$_{\pm0.03}$ & 0.89$_{\pm0.11}$ & \textbf{0.24}$_{\pm0.01}$ & 5.7 & 0.47 \\
\rowcolor{LightGreen}
InSQuaD-LD & 0.49$_{\pm0.02}$ & 0.40$_{\pm0.00}$ & 0.35$_{\pm0.02}$ & 0.67$_{\pm0.07}$ & \textbf{0.62}$_{\pm0.03}$ & 0.27$_{\pm0.02}$ & 0.40$_{\pm0.01}$ & \textbf{0.89}$_{\pm0.01}$ & 0.23$_{\pm0.03}$ & 6.1 & 0.48 \\
\rowcolor{LightGreen}
InSQuaD-GC & \textbf{0.58}$_{\pm0.03}$ & \textbf{0.43}$_{\pm0.05}$ & \textbf{0.44}$_{\pm0.02}$ & \textbf{0.68}$_{\pm0.02}$ & 0.57$_{\pm0.02}$ & \textbf{0.29}$_{\pm0.01}$ & \textbf{0.40}$_{\pm0.01}$ & 0.85$_{\pm0.03}$ & 0.23$_{\pm0.01}$ & \textbf{3.6} & \textbf{0.50} \\
\midrule
\rowcolor{lightgray}
Oracle & 0.68$_{\pm0.03}$ & 0.49$_{\pm0.02}$ & 0.64$_{\pm0.01}$ & 0.83$_{\pm0.08}$ & 0.71$_{\pm0.05}$ & 0.59$_{\pm0.04}$ & 0.53$_{\pm0.03}$ & 0.98$_{\pm0.04}$ & 0.31$_{\pm0.02}$ & 1.0 & 0.64 \\

\bottomrule
\end{tabular}

%% file: figures/ablations.tex
\begin{figure*}
  \centering
  \includegraphics[width=\textwidth]{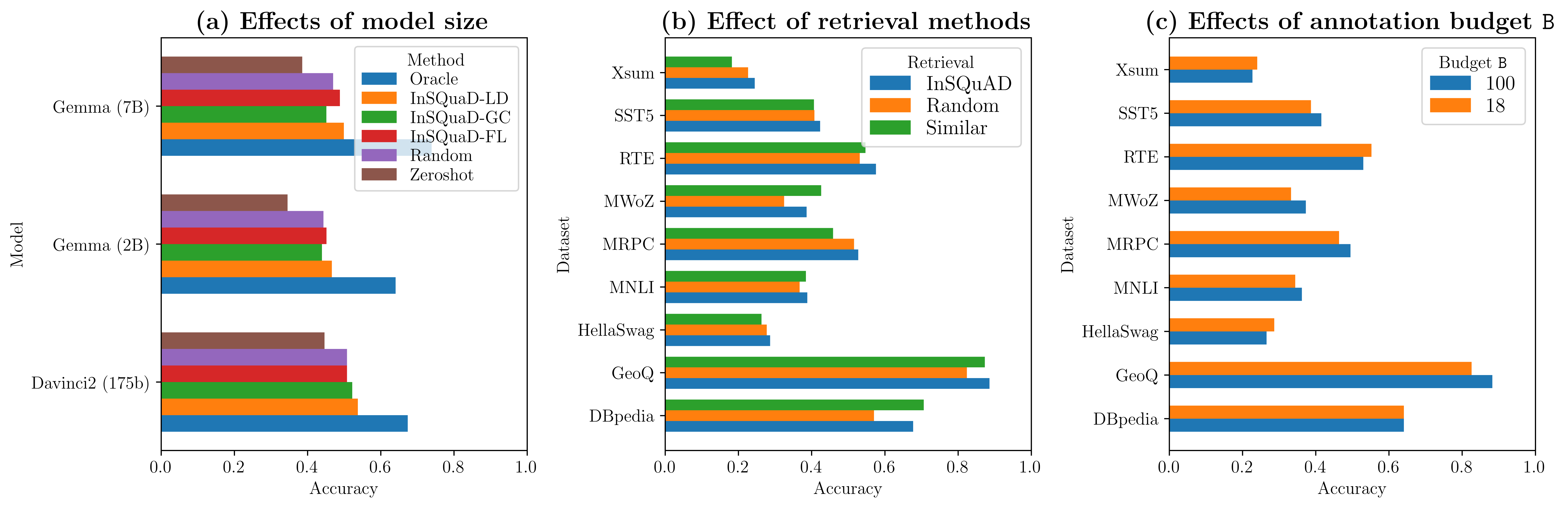}
  \caption{\textbf{Ablations in \insquad-RETRIEVE} which studies the effect of (a) Model size of the underlying LLM $\mathcal{M}$, (b) the choice of retrieval method between \{\textit{Random}, \textit{Similar}, and \textit{Combinatorial}\} with $\mathcal{M}$ fixed to gemma2b and (d) the effect of annotation budget \texttt{B} with $\mathcal{M}$ as gemma2b and $\lambda = 0.5$.}
  \label{fig:ablations}
\end{figure*}

%% file: 50_results.tex
\section{Experiments}
\label{sec:results}
\subsection{Datasets} 
We conduct our experiments on nine popular ICL benachmarks. The datasets employed in \insquad\ include \textbf{MRPC} \cite{elSaid2015ExploitingSA}, \textbf{RTE} \cite{Dagan2005ThePR}, and \textbf{MNLI} \cite{Williams2017ABC} with binary or ternary labels (yes/no/maybe). \textbf{SST5} \cite{Socher2013RecursiveDM} involves 5-way classification, \textbf{DBPedia} \cite{Lehmann2015DBpediaA} features 14-way classification, and \textbf{HellaSwag} \cite{Zellers2019HellaSwagCA} requires 4-way classification. Additionally, the generative datasets \textbf{MWoZ} \cite{Budzianowski2018MultiWOZA}, \textbf{GeoQ} \cite{Zelle1996LearningTP}, and \textbf{Xsum} \cite{Narayan2018DontGM} are employed in our experiments. 

\input{tables/hyperparameters}

\subsection{Experimental Setup}
We adopt the similar experimental setup as in \cite{Su2022SelectiveAM, Zhang2023IDEALIS}, drawing inspiration from the MetaICL framework \cite{Min2021MetaICLLT} for conducting all experiments. 
We utilize a varied set of LLMs $\mathcal{M}$ for downstream ICL tasks like Gemma (2B) \cite{Mesnard2024GemmaOM}, Gemma (7B) \cite{Mesnard2024GemmaOM}, and OpenAI-Davinci002 (175B) \cite{Brown2020LanguageMA} as the backbone models for ICL. 
We adopt the SBERT~\cite{Reimers2019SentenceBERTSE} based retrieval model $R$ in the formulation of \insquad-RETRIEVE which we train using the formulation in \insquad-LEARN to evaluate improvements in downstream tasks.

We report the hyperparameters used in our experiments for reproducibility in Table \ref{table:training_details_and_hyperparameters}. We conduct three independent trials and report results with one standard deviation confidence intervals in Table \ref{table:main_table}. All experiments use random data and weight initialization, resulting in variance even for non-trainable methods.

\begin{figure}[H]
  \centering
  \includegraphics[width=0.7\columnwidth]{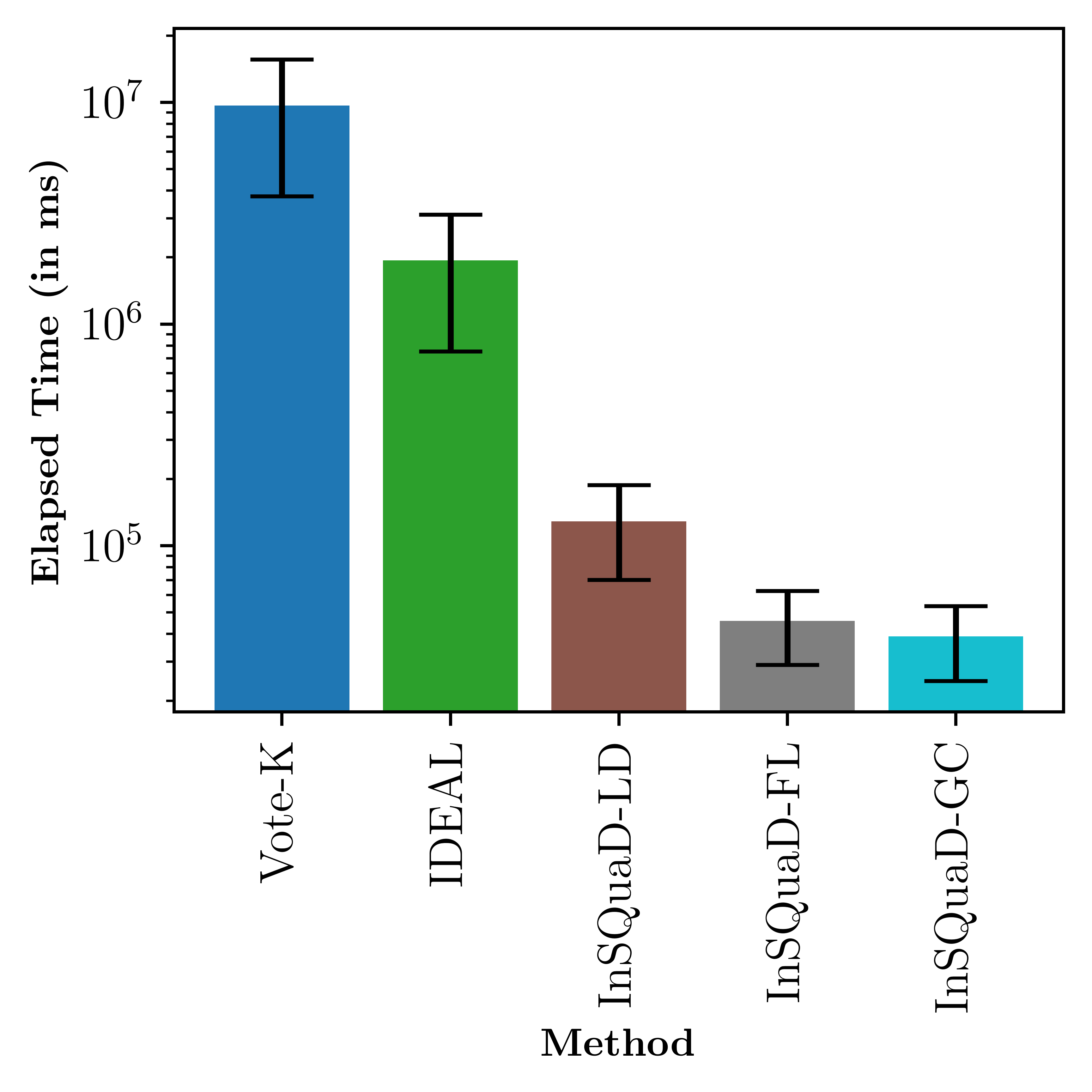}
  \caption{\textbf{Comparison of inference time} (in Log scale) across different methods. Adopting a combinatorial approach significantly reduces inference times making them a suitable choice for ICL tasks.}
  \label{fig:time_analysis}
\end{figure}

\input{tables/qd_tradeoff_gemma_ci}

The model $R$ is trained for 7 epochs. The learning rate is set at \(3 \times 10^{-5}\), a value chosen to strike an optimal balance between rapid convergence and stability in training. We adopt a weight decay of 0.01 to mitigate overfitting by penalizing large weights. The learning rate follows a linear decay strategy, gradually reducing the learning rate, which is beneficial for fine-tuning the model in its later stages. The warmup ratio is set at 0.06, allowing for a gradual ramp-up of the learning rate at the beginning of training to prevent early divergences. The optimizer of choice is AdamW \cite{Loshchilov2017DecoupledWD}, selected for its effectiveness in handling sparse gradients and adaptive learning rate capabilities.
We estimate around 527 L4 hours for all experiments. We use MPNET \cite{Song2020MPNetMA} which has 33,360,000 parameters. All experiments were conducted on a g2-standard-4 VM on GCP with 1 L4 NVidia GPU and 4 CPU cores.

\subsection{Results on Existing ICL Benchmarks}
\label{sec:main_results}
We contrast the performance of our proposed \insquad\ approach against several existing baselines tabulated in Tab.\ref{table:main_table}.
We contrast \insquad\ against Zero shot, Random selection, Vote-K \cite{Su2022SelectiveAM}, and IDEAL \cite{Zhang2023IDEALIS} among other baselines with the underlying LLM $\mathcal{M}$ as Gemma (2B)~\cite{Mesnard2024GemmaOM} and the annotation budget \texttt{B} as 18 (with additional ablation experiments in Sec.\ref{sec:quality_div_tradeoff}).

At first, we compare the performance of instances of \insquad-RETRIEVE by varying the choice of submodular function $f$ among Graph-Cut (GC), Facility-Location (FL) and Log-Determinant (LD) with pretrained model weights for the SBERT retrieval model $R$ without applying the training strategy in \insquad-LEARN. This is indicated as No Training (NT) in Tab.\ref{table:main_table}.
Our combinatorial formulation which models the exemplar selection task as a targeted selection problem shows improvements up to 6.1\% (\insquad-FL (NT)) on the multi-choice (HellaSwag) benchmark, up to 21.3\% (\insquad-GC (NT)) on the classification benchmark (MRPC), up to 8.6\% (\insquad-LD (NT)) on the Dialogue (MWoZ) benchmark and up to 24.8\% (\insquad-LD (NT)) on the generation benchmark (Xsum) over the latest baseline (IDEAL).\looseness-1

Secondly, we contrast the ICL performance of \insquad\ utilizing the SBERT based retrieval model trained using the learning formulation of \insquad-LEARN on the novel augmented dataset discussed in Sec.\ref{sec:insquad_learn}.
Similar to our earlier setting we conduct experiments on three instances of \insquad-LEARN and \insquad-RETRIEVE by varying the choice of submodular function $f$ among popular choices like FL, GC and LD.
Note, that the choice of $f$ is consistent across \insquad-LEARN and \insquad-RETRIEVE steps i.e. if $f$ is chosen to be FL, then we adopt Facility Location Mutual Information (FLMI) for targeted selection in \insquad-RETRIEVE as well as for computing $L_q$ and $L_d$ in \insquad-LEARN.
Our results show improvements of up to 21.6\% (\insquad-GC) on classification tasks, 16.4\% (\insquad-GC) on multi-choice tasks, 4.8\% (\insquad-LD) on the dialogue benchmark and up to 7\% (\insquad-FL) on generation based ICL tasks.\looseness-1

Although it is clear from Tab.\ref{table:main_table} that finetuning the retrieval model using the learning strategy of \insquad-LEARN performs significantly better than their no-training (indicated as NT) counterparts, we conduct further analysis to identify the most suitable instance of \insquad\ for ICL. Among \insquad-GC, \insquad-FL and \insquad-LD we report the average performance and average rank across all tasks (columns 11 and 12 in Tab.\ref{table:main_table}). 
From these additional insights we observe that \textbf{\insquad-GC (Graph-Cut based learning objective in \insquad-LEARN and Graph-Cut based selection in \insquad-RETRIEVE) serves as the best choice in practical settings} producing the best average performance and rank.

\subsection{Comparing Inference Time}
\label{subsec:comparing_inference_time}
In stark contrast to the iterative selection methods of \cite{Su2022SelectiveAM, Zhang2023IDEALIS}, \insquad\ employs a combinatorial approach that slashes inference times (Fig.\ref{fig:time_analysis}). Meanwhile, the confidence-based selection in Vote-k and iterative influence maximization in IDEAL inflate computational costs above \insquad's. By unifying the selection strategy with SMI functions, we significantly reduce inference times, with \insquad-GC being the swiftest. Differences among \insquad\ variants stem from kernel computations (\insquad-LD being the costliest), mirroring findings in earlier work~\cite{Kothawade2021PRISMAR}.\looseness-1

%% file: tables/hyperparameters.tex
\begin{table}[H]
\centering
\caption{Hyperparameters used in \insquad-LEARN.}
\label{table:training_details_and_hyperparameters}
\begin{tabular}{l|l}
\hline
\textbf{Hyperparameter} & \textbf{Value} \\
\hline
Epochs & 7 \\
Batch Size & 32 \\
Learning Rate & 3e-5 \\
Weight Decay & 0.01 \\
Learning Rate Decay Strategy & linear \\
Warmup Ratio & 0.06 \\
Optimizer & AdamW \\
\hline
\end{tabular}
\end{table}

%% file: tables/qd_tradeoff_gemma_ci.tex
\begin{table*}[t]
\centering
\small
\caption{\textbf{Ablation on the Quality-Diversity trade-off in \insquad-LEARN} achieved by varying the hyperparameter $\lambda$ on Gemma (2B) across tasks and learning objectives.}
\label{table:qd_tradeoff}

\setlength{\tabcolsep}{3pt}
\begin{tabular}{@{}lcccccccccc@{}}
\toprule
Method & $\lambda$ & \multicolumn{5}{c}{\textbf{Classification}} & \textbf{Multi-Choice} & \textbf{Dialogue} & \multicolumn{2}{c}{\textbf{Generation}} \\
\cmidrule(lr){3-11}
 &  & \textsc{MRPC} & \textsc{SST5} & \textsc{MNLI} & \textsc{DBpedia} & \textsc{RTE} & \textsc{HellaSwag} & \textsc{MWoZ} & \textsc{GeoQ} & \textsc{Xsum} \\
\midrule \midrule
Zeroshot & - & 0.28$_{\pm0.02}$ & 0.26$_{\pm0.02}$ & 0.41$_{\pm0.01}$ & 0.52$_{\pm0.02}$ & 0.55$_{\pm0.02}$ & 0.19$_{\pm0.01}$ & 0.07$_{\pm0.00}$ & 0.63$_{\pm0.01}$ & 0.18$_{\pm0.00}$ \\
Random & - & 0.51$_{\pm0.02}$ & 0.42$_{\pm0.01}$ & 0.37$_{\pm0.02}$ & 0.57$_{\pm0.00}$ & 0.52$_{\pm0.04}$ & 0.27$_{\pm0.01}$ & 0.27$_{\pm0.01}$ & 0.82$_{\pm0.09}$ & 0.23$_{\pm0.02}$ \\
\midrule
InSQuaD-FL & 0 & 0.50$_{\pm0.02}$ & 0.41$_{\pm0.04}$ & 0.38$_{\pm0.02}$ & 0.65$_{\pm0.01}$ & 0.56$_{\pm0.00}$ & 0.21$_{\pm0.02}$ & 0.34$_{\pm0.03}$ & 0.89$_{\pm0.11}$ & 0.23$_{\pm0.00}$ \\
InSQuaD-GC & 0 & \textbf{0.58}$_{\pm0.03}$ & 0.40$_{\pm0.01}$ & 0.37$_{\pm0.02}$ & 0.63$_{\pm0.03}$ & 0.47$_{\pm0.04}$ & 0.24$_{\pm0.02}$ & \textbf{0.40}$_{\pm0.01}$ & 0.81$_{\pm0.03}$ & 0.22$_{\pm0.02}$ \\
InSQuaD-LD & 0 & 0.46$_{\pm0.05}$ & 0.38$_{\pm0.03}$ & 0.32$_{\pm0.01}$ & 0.67$_{\pm0.07}$ & 0.53$_{\pm0.04}$ & 0.26$_{\pm0.02}$ & 0.38$_{\pm0.02}$ & 0.86$_{\pm0.07}$ & 0.23$_{\pm0.01}$ \\
\midrule
InSQuaD-FL & 0.25 & 0.38$_{\pm0.02}$ & 0.42$_{\pm0.02}$ & 0.32$_{\pm0.02}$ & 0.59$_{\pm0.04}$ & 0.53$_{\pm0.04}$ & 0.23$_{\pm0.01}$ & 0.30$_{\pm0.03}$ & 0.89$_{\pm0.04}$ & 0.16$_{\pm0.01}$ \\
InSQuaD-GC & 0.25 & 0.51$_{\pm0.01}$ & 0.40$_{\pm0.03}$ & \textbf{0.44}$_{\pm0.02}$ & 0.52$_{\pm0.02}$ & 0.54$_{\pm0.02}$ & 0.24$_{\pm0.01}$ & 0.35$_{\pm0.05}$ & 0.84$_{\pm0.03}$ & 0.23$_{\pm0.01}$ \\
InSQuaD-LD & 0.25 & 0.49$_{\pm0.02}$ & 0.40$_{\pm0.01}$ & 0.31$_{\pm0.01}$ & 0.65$_{\pm0.01}$ & 0.58$_{\pm0.01}$ & 0.25$_{\pm0.03}$ & 0.37$_{\pm0.01}$ & 0.86$_{\pm0.00}$ & 0.21$_{\pm0.02}$ \\
\midrule
InSQuaD-FL & 0.5 & 0.39$_{\pm0.04}$ & 0.41$_{\pm0.03}$ & 0.35$_{\pm0.03}$ & 0.63$_{\pm0.03}$ & 0.53$_{\pm0.00}$ & 0.27$_{\pm0.02}$ & 0.31$_{\pm0.02}$ & 0.80$_{\pm0.06}$ & 0.24$_{\pm0.01}$ \\
InSQuaD-GC & 0.5 & 0.48$_{\pm0.04}$ & 0.33$_{\pm0.02}$ & 0.33$_{\pm0.02}$ & 0.63$_{\pm0.05}$ & 0.54$_{\pm0.01}$ & \textbf{0.29}$_{\pm0.01}$ & 0.25$_{\pm0.02}$ & 0.79$_{\pm0.01}$ & 0.22$_{\pm0.03}$ \\
InSQuaD-LD & 0.5 & 0.44$_{\pm0.03}$ & 0.38$_{\pm0.03}$ & 0.35$_{\pm0.02}$ & 0.60$_{\pm0.05}$ & 0.54$_{\pm0.04}$ & 0.27$_{\pm0.02}$ & 0.40$_{\pm0.01}$ & 0.82$_{\pm0.03}$ & 0.22$_{\pm0.02}$ \\
\midrule
InSQuaD-FL & 1 & 0.42$_{\pm0.03}$ & 0.39$_{\pm0.04}$ & 0.39$_{\pm0.02}$ & 0.43$_{\pm0.02}$ & 0.51$_{\pm0.05}$ & 0.24$_{\pm0.04}$ & 0.25$_{\pm0.01}$ & 0.87$_{\pm0.01}$ & \textbf{0.24}$_{\pm0.01}$ \\
InSQuaD-GC & 1 & 0.38$_{\pm0.05}$ & \textbf{0.43}$_{\pm0.05}$ & 0.41$_{\pm0.02}$ & \textbf{0.68}$_{\pm0.02}$ & 0.57$_{\pm0.02}$ & 0.25$_{\pm0.02}$ & 0.25$_{\pm0.01}$ & 0.85$_{\pm0.03}$ & 0.21$_{\pm0.02}$ \\
InSQuaD-LD & 1 & 0.44$_{\pm0.02}$ & 0.40$_{\pm0.00}$ & 0.30$_{\pm0.02}$ & 0.67$_{\pm0.05}$ & \textbf{0.62}$_{\pm0.03}$ & 0.25$_{\pm0.02}$ & 0.37$_{\pm0.04}$ & \textbf{0.89}$_{\pm0.01}$ & 0.23$_{\pm0.03}$ \\ \midrule
\rowcolor{lightgray}
Oracle & - & 0.68$_{\pm0.03}$ & 0.49$_{\pm0.02}$ & 0.64$_{\pm0.01}$ & 0.83$_{\pm0.08}$ & 0.71$_{\pm0.05}$ & 0.59$_{\pm0.04}$ & 0.53$_{\pm0.03}$ & 0.98$_{\pm0.04}$ & 0.31$_{\pm0.02}$ \\ \bottomrule
\end{tabular}
\end{table*}

%% file: 60_ablations.tex
\section{Ablations}
\label{sec:quality_div_tradeoff} 
\noindent \textbf{Choice of retrieval method.} We examine three retrieval strategies: \textit{Random}, which assembles few-shots randomly after shortlisting; \textit{Similar}, which selects the top-$k$ most similar samples; and \insquad-RETRIEVE. We fix the training strategy for the retrieval model as \insquad-LEARN (trained using the \insquad-FL formulation) and vary the methods used for Exemplar Annotation and Retrieval. 
As shown in Figure~\ref{fig:ablations}(b), our combinatorial formulation of \insquad-RETRIEVE consistently outperforms both baselines by jointly optimizing for quality and diversity in exemplar selection. 
This experiment highlights two key findings: (1) Regardless of the retrieval method, training the SBERT model 
 with our combinatorial \insquad-LEARN objective (varied between popular submodular functions - FL, GC and LD) consistently yields notable performance gains; and (2) the combination of \insquad-LEARN (for training 
) and targeted selection via \insquad-RETRIEVE achieves the highest gains in 7 out of 9 tasks.\looseness-1

\noindent \textbf{Effect of annotation budget.} We explore the effects of varying the annotation budget $\texttt{B} \in \{18, 100\}$, using the same values as our baseline papers \cite{Zhang2023IDEALIS, Su2022SelectiveAM} for a ceteris paribus analysis. As shown in Figure~\ref{fig:ablations}(c), annotation budget does not scale with accuracy and very often does not strongly correlate with performance gains. 
This observation is also consistent with findings from our baseline papers \cite{Zhang2023IDEALIS, Su2022SelectiveAM}. 
Note, while a larger budget increases the complexity of exemplar selection by expanding the search space, our approach maintains consistent performance regardless of budget size.\looseness-1

\noindent \textbf{Effect of $\lambda$ (quality-diversity tradeoff).} We explore the trade-off between quality $L_q$ and diversity $L_d$ in $R(., \theta)$ by adjusting the hyperparameter $\lambda \in [0, 1]$. When $\lambda = 0$, the model focuses solely on query relevance (quality), ignoring paraphrase signals while, $\lambda > 0$ prioritizes both quality and diversity. 
Our results in Tab. \ref{table:qd_tradeoff} clearly portray the \textit{need for both quality and diversity in the learnt embeddings} of $R$ with best performances achieved by models with $\lambda > 0$ in 7 out of 9 tasks.
Nevertheless, its also evident that the optimal value of $\lambda$ varies based on the downstream task requiring the user to perform task specific calibration during deployment.\looseness-1

\noindent \textbf{Variation in model size.} We investigate the impact of model scale by evaluating on three underlying LLMs $\mathcal{M}(.;\hat{\theta})$: Gemma (2B), Gemma (7B)~\cite{Mesnard2024GemmaOM}, and Davinci-002 (175B) \cite{Brown2020LanguageMA}. 
Our results in Fig.\ref{fig:ablations}(a), consistent with findings of \cite{Zhang2023IDEALIS}, show that larger models show better overall ICL performance. However, we continue to use Gemma (2B) in our experiments (Tab. \ref{table:main_table}) owing to its wide adoption in latest benchmarks~\cite{Zhang2023IDEALIS, Kumari2024AnES} and low parameter counts. 
However, adopting the methodology in \insquad\ demonstrates improvements on all three model variants indicating the generalizability of our approach.\looseness-1

%% file: 70_outro.tex
\section{Conclusion, Limitations and Future Work}
\insquad\ introduces a novel combinatorial approach for In-Context Learning (ICL) leveraging SMI functions to enforce quality, diversity, and order in exemplar selection and retrieval. By pairing \insquad-RETRIEVE towards targeted exemplar selection with \insquad-LEARN for training the underlying retrieval model through likelihood-based combinatorial loss, our approach systematically improves ICL performance across nine benchmarks, validating our framework’s effectiveness.
Nevertheless, The current model adopts only the HotpotQA dataset (due to its popularity in Question Answering (QA) literature) in \insquad-LEARN leaving other multi-hop QA datasets to be experimented with as future research.
Additionally, addressing selection biases during exemplar annotation and retrieval, and improving the interpretability of our model are potential future research directions.\looseness-1  

\section*{Acknowledgements}
We gratefully thank anonymous reviewers for their valuable comments. We would also like to extend our gratitude to our fellow researchers from the CARAML lab at UT Dallas for their suggestions. This work is supported by the National Science Foundation under Grant Numbers IIS-2106937, a gift from Google Research, an Amazon Research Award, and the Adobe Data Science Research award. Any opinions, findings, and conclusions or recommendations expressed in this material are those of the authors and do not necessarily reflect the views of the National Science Foundation, Google or Adobe.